# Real-time SLAM Pipeline in Dynamics Environment

Alex Fu
Stanford University
Computer Science Department
alexfu@stanford.edu

Lingjie Kong
Stanford University
Mechanical Engineering Department
ljkong@stanford.edu

## Abstract

*Inspired by the recent success of application of dense data approach by using ORB-SLAM and RGB-D SLAM, we propose a better pipeline of real-time SLAM in dynamics environment. Different from previous SLAM which can only handle static scenes, we are presenting a solution which use RGB-D SLAM as well as YOLO real-time object detection to segment and remove dynamic scene and then construct static scene 3D. We gathered a dataset which allows us to jointly consider semantics, geometry, and physics and thus enables us to reconstruct the static scene while filtering out all dynamic objects.*

## 1. Introduction

Rapid and reliable 3D reconstruction has become a major innovation in many applications such as autonomous driving, 3D printing, virtual reality and robotics. However, fast as well as reliable dynamic scene 3D reconstruction is still under research.

Most 3D objection reconstruction methods work from case to case. Overall, there are two different major approaches. One relies on dense data SLAM to figure the best matching from frame to frame in order to reconstruct the overall 3D. The other one depends on neural networks and let neural network to learn the best matching from input Multiview images to output 3D scene. However, both cannot handle dynamic scene case.

RGB-D SLAM uses frame to frame tracking to minimize both the photometric and the depth error over pixels through global loop closure [1] [2]. However, this method requires a dense observation from Multiview. This is both computation expensive and require a huge amount of data. ORB-SLAM is another approach without utilizing pixel depth information as above. Instead, it first uses ORB extractor [3] to extract points from different frames. Best correspondence is established between two frames which has the most matched key points to form a spanning tree for all frames. Then, initial 3D points are estimated by using the spanning true frames. Last, 3D points are refined by global loop closure through using Multiview. However, this method still utilize dense data [4]. Prior knowledge with semantic priors is also integrated to monocular SLAM [5]. Known objects segmented from the sense is used to enhance the clarity, accuracy, and completeness of the map built by the dense SLAM system.

Convolutional neural networks (CNN) are widely used in image classification, localization, segmentation, and so on. Convolutional neural networks can also be used to infer a 3D representation of a previously unseen object given a single image of this object [6]. Encoder-decoder network takes RGB image and desired viewpoint as input. It output RGB and depth as the output. However, this method might require a large amount of label data for training. 3D Recurrent Reconstruction Neural Network (3D-R2N2) takes in one or more images from arbitrary viewpoints and outputs a reconstruction of the object in the form of 3D occupancy grid [7]. Even though this network does not require any image annotations or object class labels, it can only output low resolution occupancy grid.

However, none of the above approach can handle dynamic scenes. Therefore, we are proposing a dynamics 3D static scene reconstruction and dynamics scene understanding.

The specific 3D dynamic scene that we used for our project is human object interactions. There are three major aspects of our scenes: the semantics of the scene (static scene), the geometry of the scene (camera pose), and the physics of the interaction (dynamics scene).

We collect a large dataset of egocentric multi-model video of daily activities such as washing dishes and so on. We use RGB, depth, and thermal information to extract semantics, geometry, and physic information about the scene.

Even though the information from RGB, depth, and thermal images has necessary information to extract sematic, geometry, and physics of interaction, it is not structured in a clean method for segmenting out the dynamic scenes from the static ones. Therefore, we are going to introduce our method which use RGB-D SLAM pipeline and YOLO real-time object detection to remove dynamics scenes while 3D reconstruction the static scene.

Overall, our frameworks are: i) collect data from RGB, depth, thermal data which includes semantics (static scene), geometry (camera pose) and physical information (dynamics scene). ii) Use Yolo real-time detection object detection framework to detect the moving objects and remove the dynamics scene from the static scene by using energy-based minimization for segmentation via graph-cuts. iii) Use a SLAM algorithm to reconstruct the filtered 3D static scene. Eventually, we will deliver a package which will take video frames as input and output reconstructed static 3D scatter plot scenes.

## 2. Related Work

**Object Tracking:** Object detection and tracking are traditional computer vision problem. As for our case, the dynamic object that we mainly focus on is human hands in daily activities. Some hand detection algorithm relies on pixel-wise hand detection for egocentric videos combined with labeled hands data [8]. Other hand detection algorithm relies on depth-based hand pose approach [9]. We will use 3D hand pose detection algorithm which is implement under YOLO real-time object detection algorithm [10].

**Semantic Segmentation:** There are two major approaches in segmentation. One is based on using machine learning especially convolutional neural networks to learn the segmentation weights based on labelled pixels as well as category [11][12][13]. The other is based on feature extractor such as using k-mean, shift-mean, GMM, as well as energy-based minimization for segmentation via graph-cuts [14][15][16]. As for this problem, we will use energy-based graph segmentation through graph-cuts because it runs fast and do not require us to pretrain the segmentation model from machine learning approach.

**Static SLAM Reconstruction:** SLAM is a problem which contract the trajectory of object as well as the map of the environment simultaneously. RGD-D SLAM works by minimizing the RGD color error based on RGB-D camera [1] while ORB-SLAM propose a sparse, feature-based monocular SLAM system based on monocular camera [17]. Recently, LSD-SLAM proposes a direct monocular SLAM algorithm for large-scale environments [18]. ORB-SLAM2 extends from monocular cameras to stereo and RGB-D cameras. Furthermore, under the condition that explicit models of the map can increase the SLAM accuracy, we most of our scenes are recorded ego-centrically to simplify the calculation and enhance the accuracy.

However, all the proposed SLAM algorithms are designed for static scenes. Even though SLAM algorithm assumes a static environment ahead, some model are built for dynamic SLAM [19]. Another model is built based on DynamicFusion for reconstructing non-rigidly deforming scenes [20]. However, no of them handle the complicated dynamic scenes like us. Instead, most dynamic SLAM focus on single object.

## 3. Approach

The pipeline of our frameworks can be summarized in 3 stages in total: dynamical objection detection, dynamical object segmentation and filter, and static 3D scene reconstruction. The overall flow chart is shown as below.

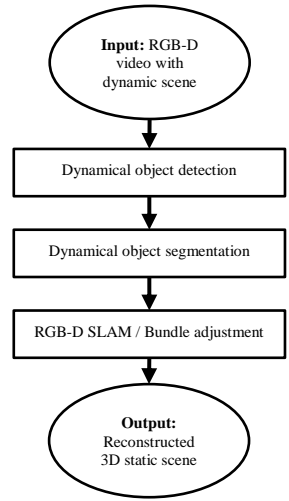

We will explain the detailed of our frameworks now.
The video is first split into RGB frames as well as depth frames as below.

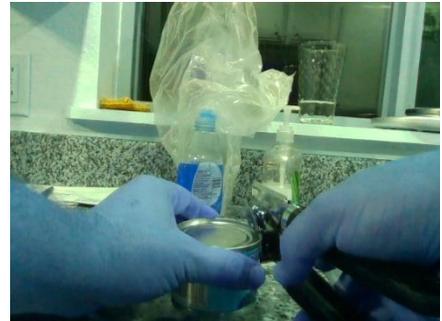

*Figure 1 RGB Frames with Moving Hand*

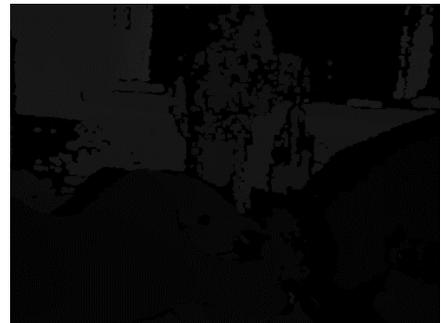

*Figure 2 Depth Frames with Moving Hand*

Each corresponding RGB frames and depth frames are tied together based on the time stamp which is clipped from the original video.

**Step 1:** First, the RGB frame is past into the object detection algorithm. The specific object detection algorithm we use from YOLO real-time object detection and classification. The details about the YOLO can be find at

[21].

YOLO apply a single neural network to the full image. This network divides the image into regions and predicted bounding boxes and probabilities for each region. These bounding boxes are weighted by the predicted probabilities. The general prediction sequence of YOLO can be summarized in the flow chart below

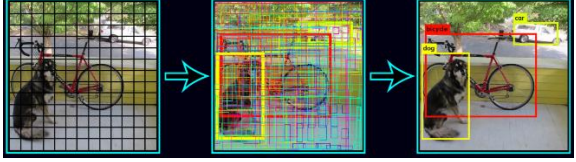

*Figure 3 YOLO Object Detection* [21]

The dataset which is used for training the object detection is 9000 classes from ImageNet as well as detection data from MS coco.

Compared to the first version YOLO which yield less localization accuracy compared to fast R-CNN, YOLO2 uses several techniques such as batch normalization, high resolution classifier, convolutional with anchor boxes and dimension cluster to outperform fast R-CNN in accuracy, while also outperform R-CNN in speed. Different from R-CNN which predicts bounding boxes using hand-picked prior, YOLO predicts the coordinate of bounding box directly using fully connected layers on top of the convolutional feature extractor as well as the class label in the bounding box.

**Step 2:** Second, after the YOLO real-time object detection algorithm detect all real-time objects as well as their classes in one frame, we know will sort out what object will be dynamic based on its class. For example, classes such as person will be more liked to move around and we will treat them as dynamics objects, while the background and other objects such as cup, bottom will be treated as static objects. We will grab the dynamic object region and then pass it into energy-based minimization for segmentation via graph-cuts.

Graph based segmentation represent features and their relationships using a graph. It cut the graph to get subgraphs with strong interior links and weaker exterior links. Eventually, the cuts between subgraphs represent the segmentation boundary.

The specific algorithm that we used is GrabCut: interactive foreground extraction using iterated graph cuts [22]. This algorithm works for foreground extraction with a bounding box feed in by the user. In our case, the bounding box covering the dynamic object is from YOLO object detection algorithm in. Eventually, it will extract the object in the bounding box from the background scene.

Then, the algorithm automatically segments the foreground region inside the rectangle bounding box from the background scene. However, in some case, the segmentation is not perfect and this will need some more user interaction. Specifically, user will need to give some strokes to mark the missing foreground as well as the contrast background. Eventually, it will show a better result which segment all foreground from the background. The detailed procedure can be summarized in the figure below.

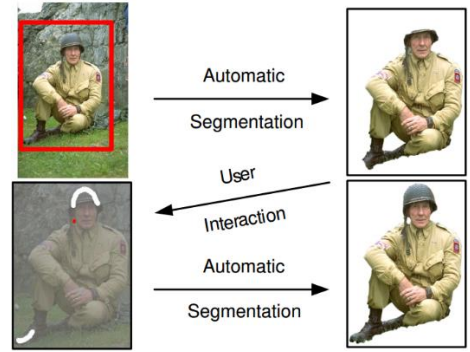

*Figure 4 GrabCut Algorithm* [22]

Now, let's see what is happening in the background in a novel version.

Let's assume that the image has $z = \{z_1, \dots, z_N\}$ RGB pixels in total. Two GMM $\theta = \sum_{i=1}^{k} \alpha_i N(\mu_i, \Sigma_i)$ will assigned for background and foreground separately. Each GMM for foreground and background will consist of $K$ components. GMM will be run until convergence based on pixel RGB color. Eventually, GMM will produce $k = \{k_1, \dots, K_N\}$ assignment for each pixel. Meanwhile, another prior label $\alpha = \{0,1\}$ will also be defined. As for foreground, $\alpha = 1$ while $\alpha = 0$ for background.

**Energy Function:** The Gibbs Energy equation will be defined as

$$E(\alpha, k, \theta, z) = U(\alpha, k, \theta, z) + V(\alpha, z)$$

**Fitting Function:** $U$ evaluate the fit of the opacity distribution $\alpha$ to the pixels z given the GMM model $\theta$

$$\begin{aligned} U(\alpha, k, \theta, z) &= \sum_n D(\alpha_n, k_n, \theta, z_n) \\ &= -\log p(z_n | \alpha_n, k_n, \theta) \\ &\quad - \log \pi(\alpha_n, k_n) \end{aligned}$$

In which $p(.)$ is the Gaussian probability distribution and $\pi(.)$ is the mixture prior weight efficient.

As for each pixel, after plugging in the mean and covariance of Gaussian distribution

$$D(\alpha_n, k_n, \theta, z_n) = -\log \pi(\alpha_n, k_n) + \frac{1}{2}\log \det \Sigma(\alpha_n, k_n)$$
$$+ \frac{1}{2}[z_n - \mu(\alpha_n, k_n)]^T \Sigma(\alpha_n, k_n)^{-1}[z_n - \mu(\alpha_n, k_n)]$$

Eventually, the GMM model $\theta$ are parametrized the parameters below that we can optimize over

$$\theta = \{\pi(\alpha, k), \mu(\alpha, k), \Sigma(\alpha, k), \alpha = 0,1, k = 1 \dots K\}$$

**Smoothness Function:** $V$ measures the smoothness

$$V(\alpha, z) = \gamma \sum_{(m,n) \in C} dis(m,n)^{-1}[\alpha_n \neq \alpha_m] \exp -\beta(z_m - z_n)^2$$

In which $[\phi]$ is the indicator function for 0 or 1, C is the set of neighbors, and $dis(.)$ is the Euclidian distance.

**Optimization:** With the energy model clearly defined, our next step is to find the best $\hat{\alpha}$ to minimize the energy, the $\hat{\alpha}$ which denotes the smallest energy will be the segmentation parameter

$$\hat{\alpha} = argmin_\alpha E(\alpha, k, \theta, z)$$

Solving the $\alpha$ analytically will be challenge. Therefore, we will use iterative energy minimization to minimize the energy step by step.

The idea is to find the min cut to resolve the energy graph as below. As for the details, one can refer to [22] GrabCut implementation as well as [23] Expectation and Maximization (EM) algorithm for Gaussian Mixed Model (GMM).

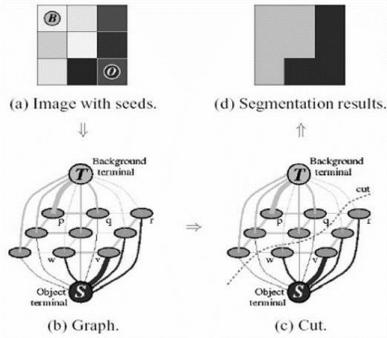

*Figure 5 GrabCut Approach* [22]

Eventually, the separation between foreground and background will be denoted by the update $\alpha$ through lots of iterations.

We used OpenCV GrabCut library to help us achieve the segmentation.

The segmented RGB image by GrabCut as well as the depth image will be past into the step 3 to achieve 3D reconstruction.

**Step 3:** This is the final step which takes in the segmented RGB images as well as the depth images to finally reconstructed the static 3D scene.

This dense visual SLAM method for RGB-D cameras work by minimizing both the photometric and the depth error over all pixels. Compared to most other feature based methods, this method allows us to better utilize the available information in the image data which yields a higher camera pose accuracy. For more details, one can refer to this paper [1].

Let's have a close look at the background.

**Camera Model:** A point in 3D homogeneous coordinate can be expressed as $P = (X, Y, Z, 1)^T$, it has corresponding pixels point $x = (x, y)^T$ and depth measurement $Z = Z(x)$. Based on the camera focus lengths $f_x, f_y$ and coordinate of the camera center $o_x, o_y$, we should be able to expressed the reconstructed 3D point by using pixel and depth value as below

$$P = \pi^{-1}(x, Z) = \left(\frac{x - o_x}{f_x}Z, \frac{y - o_y}{f_y}Z, Z, 1\right)^T$$

In which $\pi$ is the inverse projection matrix

Therefore, the forward projection matrix can be written as

$$x = \pi(P) = \left(\frac{Xf_x}{Z} + o_x, \frac{Yf_y}{Z} + o_y\right)$$

The key of RGB-D SLAM is to estimate the camera pose, in order to do so, we need to set up the rigid body transformation.

**Rigid Bode Motion:** Under homogeneous coordinate, the transformation matrix will be a 4×4 matrix including both translation and rotation.

$$T_{4\times 4} = \begin{bmatrix} R_{3\times 3} & T_{3\times 1} \\ 0 & 1 \end{bmatrix}$$

Let's define the transformation matrix function as $g$ below with P be the original point before transformation and P' be the point after transformation

$$g(P) = P' = TP$$

Obviously, 3D transformation has 6 degrees of freedom. However, matrix T has 12 degrees of freedom and gets overparametrized. Therefore, we will summarize the transformation matrix in twist coordinate $\xi$ by the Lie algebra $SE(3)$. $\xi$ is a 6 elements vector. We will define matrix exponential to calculate $T$ from $\xi$.

$$T = \exp(\xi)$$

**Wrapping Function:** Let's define a warping function $\tau$ which take in the pixel points and the depth in the first image as well as the 3D transformation to get the pixel points in its corresponding second image. The equation is shown as below

$$x' = \tau(x, T) = \pi\left(T\pi^{-1}(x, Z_1(x))\right)$$

It first reconstructs the 3D points from the first image by using the pixels and the depth information. Then, it applies the transformation matrix for the camera pose from first image to the second image. At the end, it projects the translated 3D points again into second image.

**Loss Function:** we define loss function which will evaluate the pixels and depth different from the original image as well as the projected one.

The photometric or intensity error $r_I$ for a pixel x can be defined as

$$r_I = I_2(\tau(x, T)) - I_1(x)$$

The depth error $r_z$ is defined as

$$r_z = Z_2(\tau(x, T)) - [T\pi^{-1}(x, Z_1(x))]_z$$

Last, we will minimize the overall loss for all pixels and depth across all images to find the best camera pose T.

## 4. Experiment

We first clipped corresponding RGB and depth frames from video provided by color camera as well as the depth camera. The images are shown as below.

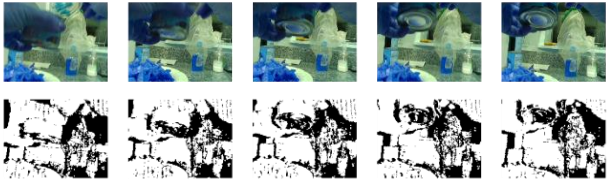

*Figure 6 RGB-D Frames 1: Can Opener*

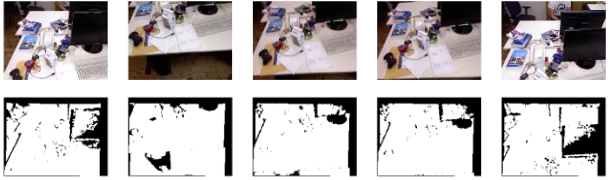

*Figure 7 Video Frame 2: Office*

We will go through on how one frame will be process below. All other frames will be processed similarly to generate the camera pose file for 3D reconstructing the 3D scene. One frame with dynamic object and the background static scene is shown as below.

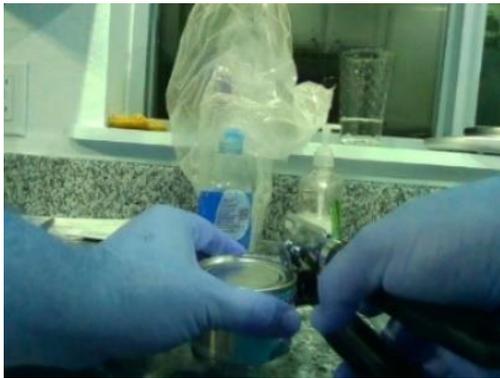

*Figure 8 1 RGB Frame Can Opener*

This frame will be past into the YOLO object detection. YOLO object detector will export a class label as well as the object location as show in the figure below

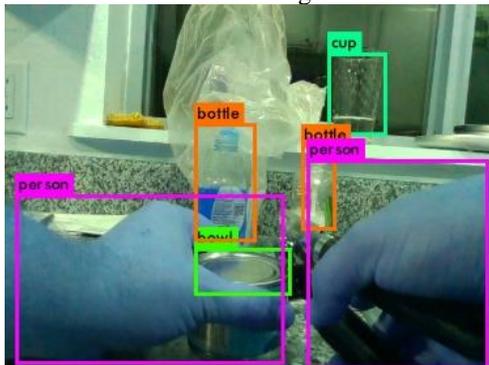

*Figure 9 YOLO Real-time Object Detection*

The confidence of each object as well as its labels can be seen as the table below

*Table 1 Object and Confidence Table*

| Category | Confidence |
| --- | --- |
| Cup | 77% |
| Bottom | 66% |
| Bowl | 47% |
| Bottle | 76% |
| Person | 37% |
| Person | 72% |

Based on the object category, we know that person will more like to be a moving object. Therefore, we grab the bounding box location of two persons and pass it to the GrabCut algorithm which to segment out the person with the prior knowledge of the bounding box as the foreground. The segmented mask can be seen as below.

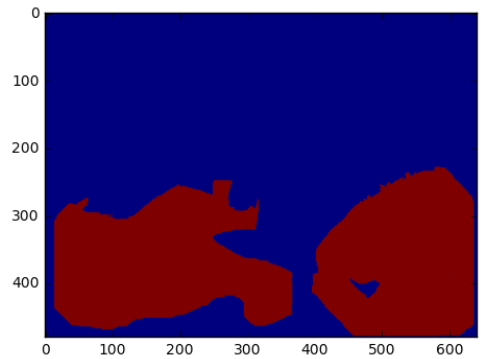

*Figure 10 GrabCut Segmentation*

We use the segmented mask to mask out the dynamic object of the RGB image as shown in the figure below and pass the filtered RGB image as well as its corresponding depth image into the last 3D reconstruction for static scene.

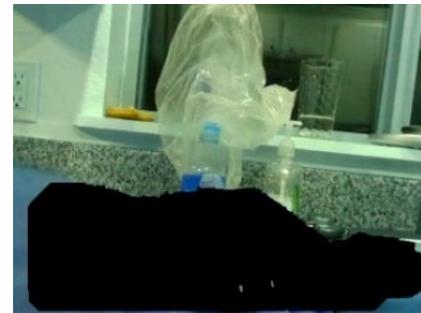

*Figure 11 Segmentation to mask RGB Image*

RGB-D SLAM will estimate the camera pose which summarize the information for transformation matrix including translation and rotation information from one camera pose to another.

The camera pose is output in the format of quaternion in order to avoid Gimbal lock. We will translate the quaternion to SE(3) translation and rotation matrix which can be used eventually for the 3D reconstruction for the static scene.

The 3D reconstruction for the first scene can opener and the second one office can be seen in the below two figures without the moving objects.

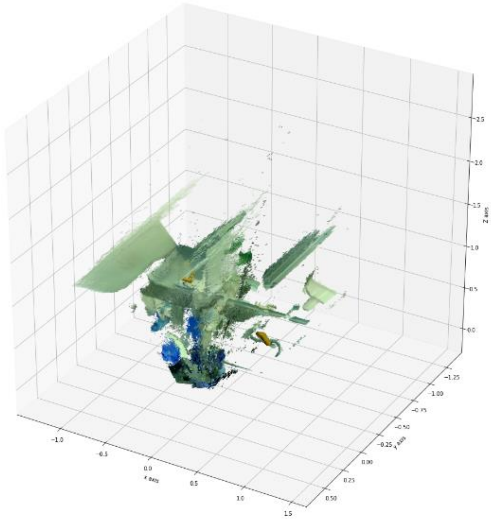

*Figure 12 3D reconstruction for Can Opener*

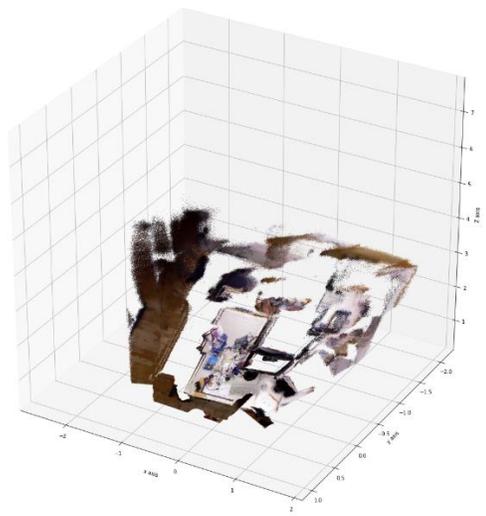

*Figure 13 3D reconstruction for Office*

However, as we can see in the first 3D reconstruction that we can still see the moving object hand. The reason is that the object detection might fail to detect the object once a while. Therefore, we might need to train a better object detector from R-CNN or YOLO to give better performance in filtering out the dynamic objects. Besides, the 3D reconstruction has successfully reconstructed the background with static scenes.

## 5. Conclusion

Recap what we have learn in this project. We have implement the real-time SLAM pipeline for 3D reconstruction under dynamic environment. The input for our pipeline is RGB-D videos frames and the output is the fully reconstructed 3D scene without the dynamic objects.

In order to do so, we first pass in the RGB images into YOLO real-time object detection framework to box the dynamic object. Second, the bounding boxes of the dynamic object will be past into the GrabCut segmentation algorithm to generate the segmentation mask which will be used to segment out the dynamic object. Third, the filtered RGB images and their corresponding depth images will be past into RGB-D SLAM algorithm to predict camera post. Eventually, the camera post will be used to generate the static 3D scene without the dynamic object.

We generate promising result to successfully reconstrued the 3D scene without most dynamic object. However, object detection might fail once a while. Therefore, we need to spend more time to training a more robust object detection algorithm. Moreover, the 3D reconstruct from dense RGB-D SLAM is time consuming and inaccurate once a while. We are considering use ORB-SLAM as well as other SLAM approaches combining bundle adjustment to generate more accurate 3D geometry as well as analytically analyze how accurate the 3D reconstruction is, compared to the ground truth.